\pdfoutput=1

\documentclass[11pt]{article}

\usepackage{acl}

\usepackage{times}
\usepackage{latexsym}

\usepackage[T1]{fontenc}
\usepackage[utf8]{inputenc}

\usepackage{microtype}

\usepackage{inconsolata}

\usepackage{graphicx}

\usepackage{times}
\usepackage{latexsym}
\usepackage{booktabs}
\usepackage{linguex}
\usepackage{enumitem}
\usepackage{natbib}
\usepackage[T1]{fontenc}
\usepackage[utf8]{inputenc}
\usepackage{microtype}
\usepackage{inconsolata}
\usepackage{graphicx}
\usepackage{amsmath}
\usepackage{tabularx}

\usepackage{listings,xcolor}

\newcommand{\new}[1]{\textcolor{black}{#1}}

\usepackage[textsize=scriptsize,textwidth=2cm]{todonotes}

\renewenvironment{quote}{%
  \list{}{%
    \rightmargin0.3cm
    \leftmargin0.3cm  
    \rightmargin\leftmargin
  }
  \item\relax
}
{\endlist}

\title{Linguistic Generalizations are not Rules: Impacts on Evaluation of LMs}

\author{Leonie Weissweiler \quad \quad \quad Kyle Mahowald \\
  The University of Texas at Austin\\
  \texttt{\{weissweiler,kyle\}@utexas.edu}\\\And 
  \quad \quad  \quad Adele E. Goldberg \\
  \quad \quad \quad Princeton University\\
  \quad  \quad \quad \texttt{adele@princeton.edu} \\   
  }

\date{}
\begin{document}
\maketitle
\begin{abstract}
Linguistic evaluations of how well LMs generalize to produce or understand language often implicitly take for granted that natural languages are generated by symbolic rules. According to this perspective, grammaticality is determined by whether sentences obey such rules. Interpretation is compositionally generated by syntactic rules operating on meaningful words. Semantic parsing maps sentences into formal logic.
Failures of LMs to
obey strict rules are presumed to reveal that LMs do not produce or understand language like humans. 
Here we suggest that LMs' failures to obey symbolic rules may be a feature rather than a bug, because natural languages are not based on neatly separable, compositional rules. Rather, new utterances are produced and understood by a combination of flexible, interrelated, and context-dependent \textit{constructions}. Considering gradient factors such as frequencies, context, and function will help us reimagine new benchmarks and analyses to probe whether and how LMs capture the rich, flexible generalizations that comprise natural languages. 
\end{abstract}

\section{Introduction}


How well do large Language Models (LMs) 
generalize beyond their training data? 
Much work on this question has presumed that generalizations require symbolic rules for syntax and semantics that generate acceptable new forms and compositional meanings. 
Rules are invoked to explain that
if you learn a new modifier (`blonky') and a new count noun (`gravimin'), a compositional rule could predict that `a blonky gravimin' is a gravimin that is blonky. 
In what follows, we use ``rule'' to refer to context-free generalizations that contain variables, to be instantiated by any instance of a general type, uninfluenced by frequency, similarity, or context \cite{pinker1999}. Our focus here is on the use of a strict algebraic conception of rules, which we argue, underlies certain approaches to NLP evaluation, even though the notion of a rule is used variably in linguistics today, with several frameworks incorporating functional and/or frequency-based attributes into representations \cite[e.g.,][]{bresnan2007predicting, brehm2022pips, odonnell2015productivity}, 


Because early statistical models (e.g., \textit{n}-gram or Markov models) seemed unable to generalize fully or capture non-local dependencies \cite{chomsky1957},  early on, rules seemed to many to be the only game in town for human language. After all, if a standard bigram model hadn't seen `blonky gravimin' before, it would be unable to form a representation of it.
Influential thinkers
argued that neural networks, which did not involve rules, would never be appropriate models of human cognition for this reason \cite{fodor1988connectionism, pinker1988, marcus1998, fodor2002compositionality, marcus2001, calvo2014architecture}.

However, current LMs arose from statistical, distributional parallel models \cite{mikolov2013a,rumelhart1986} rather than rule-based natural language technologies.
They do not rely on hard-coded rules, yet their ability to produce coherent, naturalistic language and respond appropriately is unparalleled by purely symbolic systems \cite{piantadosi2023modern, goldberg2024,weissweiler-etal-2023-construction,hofmann2024derivational}.  GPT-4o, for example, not only recognizes `a blonky gravimin' as a noun phrase, it explicitly offers several naturalistic interpretations, e.g., `A person or act that awkwardly and absurdly pretends to be serious.'

Nonetheless, an assumption that generalizations are equivalent to rules continues to motivate many evaluations of syntax, meaning, and their compositional combination: 
e.g.,  Natural Language Inference \cite{bowman-etal-2015-large}, Semantic Parsing \cite{palmer2005,reddy-etal-2017-universal}, tests of binary grammatical acceptability \cite{warstadt-etal-2019-neural, dentella2023} and rule-based compositionality \cite{kim-linzen-2020-cogs}. Together, such tasks made up more than half of the GLUE benchmark \cite{wang-etal-2018-glue}, created to evaluate language models on their skill at being ``general, flexible, and robust.'' Lackluster performance on rule-based tasks in the early days of LMs was taken to imply that the models did not use language the way people do and were instead merely imitating shallow surface patterns \cite{bender-koller-2020-climbing, kim-linzen-2020-cogs, weissenhorn-etal-2022-compositional,bolhuis2023}. 
In a survey of 79 NLP researchers, \citet{mccurdy-etal-2024-toward} reported that 87\% believed LMs were not sufficiently compositional and a sizable proportion (39\%) believed explicit discrete symbolic rules were required.

Evaluations of  LMs' early challenges with algebraic or logical rules did expose certain shortcomings in their ability to reason abstractly and solve math problems \cite[see e.g., ][]{Mahowald2024Dissociating}. At the same time, LM's concurrent ability to produce and respond to natural languages \textit{naturalistically} is hard to overstate \cite[e.g.,][]{coil-shwartz-2023-chocolate}.

Mastering a natural language requires mastering a network of hundreds of thousands of context-dependent, gradient, flexible schemata (\textit{constructions}, see §\ref{sec:cxns}), which often contain `slots' that constrain their fillers and how those fillers are interpreted.
Constrained slots allow for new combinations, flexibly adapted in context. For instance, the phrase '<time period> ago' can coerce a temporal interpretation of filler phrases that do not designate time periods (e.g., `three rest stops ago'). Rather than rule-based compositionality,  composition-by-construction allows constructions to contribute meaningfully to interpretation in ways that range from abstract to quite narrow and specific. Therefore, for LMs to use language like humans, they require interpretations that are far richer than rules can provide for thousands of collocations, conventional metaphors, idioms, and context-dependent interpretations. Even abstract grammatical patterns also regularly convey semantic and/or pragmatic information that restrict their contexts of use and interpretations. 
Since different languages and dialects provide speakers with different networks of constructions (ConstructionNets),  cross-linguistic differences can be captured naturally.



 We suggest that rule-based evaluations have been over-emphasized in the domain of natural language production and comprehension. Our goal is to emphasize the importance of recognizing context, frequencies, meaning and other gradient functional factors in modern evaluations of natural language. 
 
We do not argue that no categorical rule exists in any language. 
If a categorical rule is needed, it can be treated as the limiting case, a fully abstract construction \cite{jackendoff2002}.  For instance, \citet{jackendoff2002} proposes a symbolic Verb + Particle rule for the syntax of English complex verbs. At the same time, the meanings of individual verb plus particle combinations are far from compositional by any general rule (e.g., one can \textit{look up} a number or \textit{look down on someone} but not \textit{?look up on someone} nor {\textit{?look down a number}).
Here we advocate for an increased focus on the extent to which and \textit{how} LMs manage to produce and comprehend human-like natural languages in all their context-specificity and complexity.

Many of our theoretical points are not new, particularly in the domain of morphology. Neural network researchers have continuously argued in favor of a single representational system and against the usefulness of rules in the domain of words and inflectional morphology \cite[e.g.,][]{rumelhart1986, rogers2004, elman2009, christiansen1999, macdonald1994, mcclelland2015capturing}. While early work in Artificial Intelligence relied on algebraic rules \cite{minsky1969, lenat1995}, many researchers soon realized that rules were too brittle to scale up beyond highly restricted domains such as artificial block worlds \cite{winograd1980}.

\new{Our contribution is to review leading paradigms used in LM evaluations of syntax (§\ref{sec:syntax}), semantics (§\ref{sec:semantics}), and compositionality (§\ref{sec:compositionality}). 
We argue that, while these paradigms have been fruitful, they inherit from a tradition that was overly focused on rules, hierarchy, compositionality, and a binary notion of grammaticality. 
We briefly characterize how these assumptions arose and how they are baked into evaluations.  
We argue that \textbf{evaluations that presume categorical and strictly compositional language ignore some of the richest elements of human language}.
We review construction-based and gradient functionalist approaches to language, arguing that this tradition points to certain lacunae in existing evaluations and open up new possibilities for evaluating natural language understanding. Early work in this direction has already included more nuanced metrics, measuring gradient judgments and context-dependent interpretations \cite[e.g.,][]{juzek-2024-syntactic,hu2024language}.}

\section{Syntactic Rules in LM Evaluations}
\label{sec:syntax}

\new{Evaluations of the syntactic capabilities of LMs have frequently assumed a binary categorical notion of grammaticality, which is then used to create datasets for evaluation. Below, we discuss several such cases, attempting to make these assumptions explicit to show their limitations.}


\paragraph{Grammaticality Judgment Tasks}

Human judgments on sentences are gradient rather than binary, and demonstrably depend on frequency, plausibility, complexity, memory demands, potential alternatives, and context \cite{grodner2005,  schutze2013judgment, robenalt2015, gibson1993,fang2023}.  
The amount of exposure to written language or linguistic theory also influences people's judgments. For instance, \citet{dabrowska2010} found that laypeople's judgments on sentences containing long-distance dependencies were more sensitive to lexical content than linguists' judgments were. Even sentences included in linguistic textbooks, which one might presume to have clear-cut judgments, in reality are judged gradiently by people \cite{juzek-2024-syntactic}.
Nonetheless, LMs' language skills are often evaluated on binary grammaticality judgments on sentences \cite{dentella2024,dentella2023,warstadt-etal-2019-neural}.

The fact that human judgments are gradient can have profound consequences on evaluations. For instance,
\citet{dentella2023} compared humans and LMs against predetermined binary acceptability labels, reporting that LMs' performance correlated poorly. 
However, comparing gradient perplexity measures with the same human judgments
revealed a strong positive correlation \cite{hu2024language}. 
Using perplexity measures for models (as well as allowing humans to provide ordinal or gradient judgments) is a step in the right direction \cite{hu2024language, juzek-2024-syntactic}.

\paragraph{Dependency Parsing as Evaluation}
Parsing text for universal dependencies \cite[UD,][]{de-marneffe-etal-2021-universal} has become a well-established task for evaluating models \cite{zeman-etal-2017-conll,zeman-etal-2018-conll}, and since \citet{hewitt-manning-2019-structural} showed BERT \cite{devlin-etal-2019-bert} to be somewhat skilled in UD, it has became the default operationalization of syntax in the NLP world \cite{amini-etal-2023-naturalistic,kryvosheieva-levy-2025-controlled,muller-eberstein-etal-2022-probing} and in discussions of inductive biases \cite{lindemann-etal-2024-strengthening,glavas-vulic-2021-supervised}. 
UD annotations are partially determined by semantics and they are based on lexical items, which makes them closer to the 
approach advocated here rather than abstract phrase structure rules. However, UD analyses presume a universal set of grammatical relations, which is problematic, because not all languages employ the same constructs. That is, there is no universally valid way to define or test for the syntax of nouns, verbs, adjectives, subjects, or direct objects \cite[e.g.,][]{croft2001radical}. Moreover, UD requires an asymmetric relationship between a 'head' and dependent, yet the long tail of language includes headless constructions (e.g., \textit{the Xer, the Yer} construction) \cite{michaelis2003headless} and co-headed constructions (e.g., phrasal verbs, conjunctions, idioms). Therefore, UD annotations need to be determined for individual languages and need to allow for non-headed or co-headed cases to align well with formal aspects of natural languages. 

\section{Semantic Rules in LM Evaluation}
\label{sec:semantics}


Formal logic was developed as a branch of mathematics, used to prove mathematical and philosophical theorems and identify provability gaps \cite{frege1918,russell1905,goedel1931}. It was based on algebraic rules operating on categorical and broadly defined categories.
Notably, many logicians did not generally assume nor endorse using formal logic to represent the meanings of natural language utterances \cite{carnap1937,baker1986language}, recognizing that
natural languages differ from formal logic in many ways.\footnote{\new{Some logicians did advocate for using formal logic for natural language  \cite{tarski1944, montague1970,Montague1973}.}} For instance, logic treats \textit{and} and \textit{but} as equivalent. It does not provide a natural way to capture commands or questions \citep{austin1975}, nor does it naturally distinguish presuppositions from assertions \cite{strawson1967}. Finally, formal logic is not intended to capture effects of context \cite{wittgenstein1953,russin2024}. 

\new{Yet the assumption that natural language semantics can be modeled by formal logic has been made in the design of certain classic LM understanding benchmarks. Below, we review some instances and discuss their connection to formal semantics.}

\paragraph{Natural Language Inference}
Natural Language Inference tasks label the second of two sentences as an entailment, contradiction, or neutral, and this NLI task was originally used to train models \cite{superglue,dagan2005,nie-etal-2020-adversarial}. Today, NLI is used as a zero-shot evaluation metric to assess natural language understanding  \cite{zhou-etal-2024-constructions,mccoy-etal-2019-right}. In introducing the Stanford NLI corpus, \citet{bowman-etal-2015-large} state, ``The semantic concepts of entailment and contradiction are central to \textit{all} aspects of natural language meaning. '') \cite[emphasis added, see also][]{katz1972,vanBenthem2008}.
While in the same paper, \citet{bowman-etal-2015-large} acknowledge that judgments depend on many factors, such as commonsense knowledge, this fact is generally overlooked in papers that use NLI as a task to evaluate LMs' general understanding. 

 Necessary and plausible inferences \textit{are} a critical aspect of natural language understanding. However, they are highly dependent on the interlocutors' general communicative goals. We aim to make sense of others' messages, so we assume others are trying to be relevant and helpful and do our best to assign coherent meanings to all utterances \cite{grice1975}. For example, outside of logic classes or heated arguments, people rarely conclude that two statements made by the same person are contradictory. If someone utters: `The boy is depressed but he is not \textsc{DEPRESSED}', listeners do not throw up their hands and shout `contradiction!'. Instead, they may infer that the boy in question is only somewhat, and not extremely, depressed, or ask to learn more. People also assign interpretations to statements in ways that differ from what formal logic would predict (e.g.,  `run fast and you've got this' or `If it snows, it snows.'  NLI tasks that rely on judging contradictions or entailments may over- or under-estimate how well LMs' understanding of natural language aligns with humans', particularly when binary judgments are required \cite[cf.][]{dentella2024}.

Evaluation metrics need to take humans' communicative goals into account and allow for gradient and context-dependent interpretations. An example of the type of evaluation we endorse can be found in the underappreciated NOPE testbed. \citet{parrish-etal-2021-nope} selected 10 distinct constructions that trigger presuppositions and curated 100+ instances of each one, based on naturally occurring examples. Each stimulus includes two preceding sentences for context. The authors then collected gradient judgments from human raters, allowing them to use their `background knowledge about how the world works' and compared the accuracy of several models, with several controls in place. This strikes us as a highly valuable blueprint for modern evaluations of LMs.

\paragraph{Semantic Parsing}
\citet{banarescu-etal-2013-abstract} introduced abstract graphical meaning representations (AMR) for sentential meaning that importantly includes aspects of lexical semantics. It was created to offer a repository of structured meanings to be used for evaluating understanding in LMs \cite[][see also §\ref{sec:compositionality}]{li-etal-2023-slog,qiu-etal-2022-improving,shaw-etal-2021-compositional}. Yet work on AMR concedes that it ignores so-called `syntactic idiosyncracies.' For example, `he described her as a genius' and `his description of her: genius' are assigned the same AMR. Yet the former is unambiguously about her intellect, while the latter may instead be used to compliment \textit{his cleverly tactful description}.
More generally, simplifications of distinctions made in a natural language can be expected to result in lost meaning, since two utterances are rarely interchangeable in all contexts.  \new{Focusing on  subtle but important differences in meaning offers an opportunity to design more challenging linguistic evaluations of LMs.}

\section{Compositionality}
\label{sec:compositionality}

As computer coding languages became more and more widespread, rule-based syntax and semantics took root in linguistics. A Principle of Compositionality states that the semantics of a sentence is determined by the meanings of the words and the syntactic rules used to combine them \cite{montague1970, partee1984nominal, dowty1979, jackendoff1992semantic, fodor2002compositionality}. It is intended to be a bottom-up process: syntactic rules combine words, which have determinant meanings. 
\citet{fodor1988connectionism} make this clear:  ``a lexical item must make approximately the same
semantic contribution to each expression in which it occurs''. That is, context may not influence the interpretation of words in a top-down manner; therefore downstream inferences are required to address the fact that interpretations \textit{do} depend on context. Realizing this, like Carnap and Frege before him, \citet{fodor1988connectionism} acknowledge: ``It’s uncertain exactly how compositional natural
languages actually are.'' 

Nonetheless, compositionality is often taken as a truism, based on the standard argument for it summarized below.
\begin{quote}
People tend to agree on the interpretation of new sentences. 
$\Rightarrow$ There must be some set of rules that determine the meaning of new sentences. \label{rules_exist}
\end{quote}

\noindent Note that one can agree with the premise without accepting the consequent. 
In particular, people generally agree on the interpretations of pointing gestures and novel words as well as sentences, and yet shared interpretations \textit{must} be gleaned from non-linguistic context in the case of pointing gestures, and from a combination of linguistic and non-linguistic context in the case of novel words. Shared interpretation of sentences likewise comes in part from linguistic and non-linguistic context.

Consider the sentence, `the Persian cat is on the mat.' If the speaker's goal is simply to help someone find the furball, there need be no commitment to the cat being a thoroughbred Persian breed nor to the cat being wholly on, rather than adjacent to, the mat. Or, comprehenders may appreciate the statement is ironic if the cat is hairless. 

Cases that might seem amenable to a rule often turn out to require a good deal of item-specific memory. For instance, a compositional rule involving set-intersection may seem appealing for `<color term> noun' combinations in the domain of artificial block worlds (e.g., a green cube is something that is both a cube and green). However, violations of such rules abound: green tea is more yellow than green, and Cambridge blue is actually green. Even more common are instances that evoke richer meanings than predicted by any algebraic rule: e.g.,  a green light implies that forward motion or progress is permitted, and a green card provides a path toward citizenship in the US (or ought to). The meanings of familiar collocations are typically not fully determined by general compositional rules, and novel cases can be interpreted on analogy to familiar cases rather than according to some very general rule. For instance, if "flam" is interpreted to mean any kind of event or action, \textit{a green flam} is likely to be interpreted to imply an eco-friendly or beginning-level event. \new{Representing only the rule-compliant cases in evaluations can therefore lead to the wrong conclusions. A more comprehensive evaluation paradigm should take into account how people actually interpret familiar and novel cases.}

Another issue is that rules massively over-genenerate. That is, rules predict all manner of odd locutions \cite{pawley1983,sag2002}: e.g.,  `Meeting you is pleasing to me'; `The tall winds hit the afraid boy'; `Explain them the problem.' Humans are sensitive to the frequencies of various types of word combinations and judge formulations unnatural if there exists a more conventional way to express the intended message in context \cite[e.g.,][]{goldberg2019explain}.

\paragraph{Evaluating LMs for Compositionality}
Compositionality benchmarks combine elements from syntactic and semantic evaluations. %
 \citet{kim-linzen-2020-cogs}'s compositional generalization challenge (COGS) tested whether models could translate any sentence generated by a small set of syntactic rules into formal semantics. For instance, trained on representations of `the girl,' `the cat,' `the hedgehog,' `the cat loves the girl,' `the hedgehog sees the cat,' and so on, the model was tested on how well it predicted a formal semantic representation of `The girl loves the hedgehog.'  However, note that if `mosquitoes' is substituted for `the cat,'  different interpretations of `love' are evoked (`Mosquitoes love the girl' vs. `The girl loves mosquitoes'), not to mention different degrees of plausibility.
The authors also anticipated generalizations from sentences like `Jane gave the cake to John' to `Jane gave John the cake,'  and the models were found to perform poorly. Yet the two sentences differ in terms of information structure \cite{bresnan2010} and the relative frequencies and similarities of verbs witnessed in each version \cite{leong2024, ambridge2014, hawkins-etal-2020-investigating}. \new{Thus, while an evaluation of this kind can capture something about how humans interpret automatically generated sentences in an experimental context, focusing on this type of task may distort our view of how well LMs handle natural language in the wild.}

Other compositionality benchmarks adopt NLI tasks, which commonly presume interpretation is determined by rules.  
For example, in the context of robotic agents interpreting instructions, \citet[p.1]{lake2018} state:
\begin{quote}
Humans can understand and produce new utterances effortlessly, thanks to their compositional skills. Once a person learns the meaning of a new verb `dax', he or she can immediately understand the meaning of `dax twice'... 
\end{quote}
The robotic agents struggled to interpret the rule-based command, though it was appropriate in the narrow domain tested. Notably, the rule does not apply to natural language generally. For instance, unbounded actions are not countable, so if `twice' appears at all, it is likely followed by a comparative phrase (e.g., `work twice as hard'), which has a very different meaning than performing an action two times. Other cases require knowledge of specific combinations: `to think twice' means 'to hesitate' and `going twice' tends to evoke the context of an auction. Familiar phrases with meanings not fully captured by compositional rules are common: By one estimate, we learn tens of thousands of them \cite{jackendoff2002}. 
Importantly, we tend to agree on their interpretations, even though each means something more or different than predicted simply by the words and their syntactic combination.  In this way, phrasal combinations regularly involve subregularities or item-specific interpretations not predicted by general algebraic rules.

Another example comes from the seemingly innocuous algebraic rule:
``If X is more Y than Z, then Z is less Y than Z, irrespective of the specific meanings of X, Y, and Z'' \cite[:5]{dasgupta2020}.
This is meant to capture that `Anne is more cheerful than Bob' should both contradict `Anne is less cheerful than Bob', and entail `Bob is less cheerful than Anne.' NLI models that failed to draw these inferences were considered lacking. 
Yet natural language rarely relies on free variables. The content of X, Y, and Z matters.  No one would infer that because Anne$_{x}$ is more cheerful$_{y}$ than careful$_{z}$, that `Careful$_{z}$ is less cheerful$_{y}$ than Anne$_{x}$.' Perhaps more importantly, if a speaker uttered `Anne is higher than Bob and Bob is higher than Anne,' listeners would likely infer either that Bob climbed above Anne in the time it took to utter the first clause or that Bob has been smoking. 
\new{We have so far argued that an overemphasis on symbolic abstract rules for natural languages can lead to evaluations of natural language that are not aligned with humans. Below we suggest an alternative approach to language, which we argue helps refocus evaluations on interesting new research questions}.

\section{The Constructionist Approach}
\label{sec:cxns}

\new{This section briefly explains the constructionist approach to language, which conceives of a language as a vast network of interrelated \textit{constructions}, of varying size and complexity.   This differs from a perspective that treats languages as a set of sentences generated by a small set of algebraic rules. We suggest a change of perspective about the nature of language, not a mere substitution of the units on which some type of rules operate.}
\new{
That is, certain traditional evaluations were far too limited in requiring models to adhere to strict compositionality, when humans do not. At the same time, the constructionist approach encourages stringent evaluations by testing whether models capture the gradient and function-sensitive patterns that characterize natural languages.}\footnote{For more comprehensive introductions \new{to the constructionist approach, or `Construction Grammar'}, see \citet{hoffmann2013oxford} and 
\citet{hoffmann2022}.} The approach encourages us to broaden our view of language and linguistic evaluations of LMs.


\paragraph{(Partially-filled) Words, Common \& Rare Schemata as the same type of Representations} 
A `construction' is any learned association between a formal pattern and a range of related functions. This simple definition treats words, idioms, rare \textit{and} common grammatical patterns as constructions. \new{As a result, the lexicon and syntax are not treated as distinct or modular systems. This allows the many parallels between them to be easily captured. It also allows a natural way to allow for the diversity found in the world's languages, in which more or less information is encoded in a single word.} Formal attributes of constructions include phonology, grammatical categories, word order, discontinuous elements, specific words or morphemes, and/or intonation.  Any construction may include one or more constrained open `slots'.


\paragraph{A Wide Range of Functions Considered Jointly with the Forms} 
Constructions' functions vary widely: 
words, collocations, and idioms convey rich, specific, contentful meaning. A plethora of other constructions are productive but constrained in a variety of semi-specific ways; argument structure constructions convey `who did what to whom'; discourse structuring constructions indicate which parts of an utterance are at-issue or backgrounded. A range of constructions exist to ask questions, express surprise or disapproval, greetings or gossip. Construction can be associated with specific registers, genres, and/or dialects. 
\new{The constructionist commitment to considering semantics jointly with syntax represents a more comprehensive understanding of their interactions, which can help develop tests that evaluate both.}


\paragraph{Sensitivity to Similarity and Frequency}
Language users are sensitive to the frequencies of constructions. For instance, the passive construction is used far more frequently in Turkish than English and young Turkish speakers learn the construction far earlier than English-speaking children \cite{slobin1986}.
Constructions are also influenced by similarity: Instances of a construction prime instances of the same or closely related construction \cite[e.g.,][]{dubois2014, pickering2008}. 
\new{Constructionist approaches take this to be a core aspect of language and language learning, rather than an inconvenience or afterthought. This leads to a de-emphasis of definitional boundaries and an organic incorporation of fuzzy boundaries and prototypicality effects.}

\paragraph{Productive Constructions May Include Fixed Lexical Units}
Syntax, semantics, and morphology are interrelated rather than assigned distinct levels. This is useful because even productive hierarchical constructions often include particular words and semantic constraints. For example, an English construction that implies real or metaphorical motion allows a wide range of verbs but requires the particular noun `way' (`He charmed his way into the meeting.').

\paragraph{Interrelated Network, Not an Unstructured Set}
Unlike rules, which are commonly presented as unstructured lists, constructions comprise a network of interrelated patterns. This allows for the fact that each language includes families of related constructions. It also allows for the simple fact that productive constructions simultaneously co-exist with specific conventional instances. For instance, the English `double object' construction is productive, and speakers are also familiar with dozens of conventional instances  (e.g., `give <someone> the time of day', `throw <someone> a bone'). 

\paragraph{More Maximal than Minimal}
\new{A ConstructionNet includes words as well as grammatical patterns, and lossy instances are included as well as generalizations across instances, as just mentioned, which provides some redundancy. There is no reason to restrict the complexity of constructions or their descriptions more than is warranted by psychological and linguistic evidence.}

\paragraph{Construction Slots Are Constrained}
The open `slots' of constructions are constrained in a wide variety of ways. For instance, the English double-object construction can appear with a wide range of verbs, but prefers simple verbs to those that sound Latinate (e.g., `She told them something' vs. `She proclaimed them something'). The English comparative suffix `-er' (e.g., `calmer', `quicker') is available for most single-syllable adjectives that allow a gradient interpretation, but it is not used with past participle adjectives (? `benter').

\paragraph{An Example} Consider `X is the new Y'. It is productive and can be used to create new utterances, e.g.,  `Semiconductor chips are the new oil.' As is typical of productive constructions, the generalization co-exists with several familiar instances (e.g., `50 is the new 40'; `Orange is the new black'). The construction is not an algebraic rule: Its slots, indicated by X and Y, are not variables that range freely over fixed syntactic categories. Instead, `X' must be construed (playfully) as currently functioning in the culture as `Y' used to. Therefore, not all combinations of slot fillers make sense:  (e.g., ? `Orange is the new oil'). Adding a parallelism constraint between X and Y is insufficient since `103 is the new 101' would also require an unusual context to make sense.
Finally, instances of the construction are not amenable to a general compositional rule, nor can they be translated into formal logic. Either approach would presumably treat `Orange is the new black' as equivalent to `Black is the old orange,' which does not conventionally evoke the same meaning.

\section{Implications Beyond Natural Language}
\label{sec:beyond}

Outside of natural language, even in domains that are rule-like by design, rule-based interpretations are sometimes lacking, potentially due to the fact that natural language is used by people when discussing these domains.  
For instance, LMs have been found unreliable at drawing the following inference, which the authors dubbed the \textit{reversal curse}: 
``if `A is B' [...] is true, then `B is A' follows by the symmetry property of the identity relation" \cite[p. 2]{berglund2023reversal}.

\textit{Why} are LMs prone to the reversal curse?  
Although the quote above is stated in natural language, it does not apply to natural language sentences, which are rarely reversible without a different interpretation: e.g.,  `A mental illness is the same as a physical illness' means something very different than  `A physical illness is the same as a mental illness' \cite[see also][]{tversky1977features, talmy1975}.
Even simple conjunctions are not generally reversible in natural language. For instance, `night \& day' and `day \& night' are both acceptable, but their interpretations differ:  the former conveys a stark contrast (e.g., `as different as night and day'), the latter suggests a relentless activity or process (e.g., `he worried day and night').  In summary, it is perhaps reasonable to expect truly symmetric knowledge to be reversible. But LMs are trained on natural language, which is not symmetric.

\section{New Directions for Evaluation}
\label{sec:new}

Natural languages involve complex and context-sensitive systems of constructions, which vary from being wholly fixed to highly abstract and productive. Constructions are combined when a unit, potentially itself composed of constructions, fills a slot in another construction.
Viewing language as a system of constructions rather than words and rules may fundamentally change how the successes and failures of models are construed, and new goals and questions come into focus. \new{The complexity of constructions with respect to gradience in frequencies, functions, slot constraints, and prototypicality can be used to develop evaluations that demand the same complexity from LMs found in natural languages.}

\new{A caveat is required for low-resource languages, where rule-based linguistic evaluations \cite[e.g.,][]{jumelet2025} can be useful. More generally, evaluations should meet models where they are: if the representational complexity of an LM is restricted, restricted types of evaluations are required. But when evaluating LMs on high-resource languages, richer evaluations are appropriate. }
Specifically, we recommend the following.

\textbf{When possible, use a variety of naturalistic sentences} rather than sentences generated by a template that presupposes grammatical rules with interchangeable vocabulary items, as is done, e.g., by Multi-NLI \cite{williams-etal-2018-broad}. The idea that sentences can be constructed by subbing random lexical items into templates often misses lexical subtleties that are an important part of natural language.
Instead, ecologically valid stimuli
can be collected or adapted from natural corpora and normed for naturalness and plausibility. Since human judgments are highly context-dependent, benchmark tasks should also vary contexts systematically \cite[see, e.g.,][]{ross-etal-2024-artificial,parrish-etal-2021-nope}.

\textbf{Collect human assessments that allow for gradient context-sensitive interpretations that appeal to learned constructions.} Evaluating LM competence on individual constructions requires assessing both acceptability judgments and interpretations from humans to draw appropriate comparisons.

We also need to be sensitive to the implications people and LMs draw from instructions and the testing context. \textbf{Do not give instructions to human-evaluators (or models) in ways that make the results a foregone conclusion.} If people are instructed to interpret `red X' as `X that is red for any X,' they are capable of doing so; this may reflect the instructions, not their natural intuitions. In natural contexts, people understand that red grapefruits are closer to pink, red hair is more orange, a red book may be about communism, and crossing a red line may have consequences.

The items included in testing also influence interpretations by people and models, by providing context. For instance, if certain pairs of items are mismatched (e.g., ``The cup is green.'' ``The cup is blue.'') while others are matched, people can infer which ones are intended to be contradictory. LMs, like people, are now capable of generalizing by rule when tasked to do so. For instance, \citet{lampinen2025} found Gemini 1.5 Flash \cite{gemini15} avoided the reversal curse, achieving 100\% accuracy, when the \citet{berglund2023reversal} dataset was provided to the LM as context.

\textbf{A variety of items, participants, and contexts ought to be valued as much as a variety of models.}
It has long been recognized that real words, phrases and sentences vary in an open-ended number of ways \cite{clark1973}. So care must be taken to include a variety of stimuli items. Because linguistic meaning is deeply tied to local context, even seemingly similar sentences can have very different interpretations in ways that depend on context. Different subgroups of participants may perform differently so distinct dialects should be taken into account.

\vspace{12pt}
The most interesting questions may no longer be \textit{whether} LMs are skilled at producing and responding to natural languages, but \textit{how} they achieve such remarkable skills. As is familiar from the lexicon, constructions comprise an interrelated network. We can now \textbf{how relationships between constructions are picked up by LMs.}
For instance, \citet{misra-mahowald-2024-language} have demonstrated that even when all instances of a rare non-compositional construction are ablated from training data, non-trivial learning of the construction remains, enabled by the presence of related constructions in training. Moreover, nearly every productive construction coexists with at least a few formulaic instances, and LMs offer ways to test various theoretical perspectives on the nature of those relationships. These and other newer ways of probing LMs are possible, and our toolkit will only grow. 

\new{As discussed by \citet{weissweiler-etal-2023-construction}, investigating whether LMs distinguish subtle meaningful differences between constructions is another important direction. Recent work on this has included \citet{weissweiler-etal-2022-better}, who found LMs reliably discriminated instances of the English Comparative Correlative from superficially similar expressions. \citet{tayyar-madabushi-etal-2020-cxgbert} tested a dataset of automatically induced constructions and reported that BERT \cite{devlin-etal-2019-bert} could determine whether two sentences contained instances of the same construction. As mentioned earlier,
\citet{tseng-etal-2022-cxlm} showed that LMs gradiently predict appropriate slot fillers. 
\citet{li-etal-2022-neural} probed RoBERTa's implicit semantic representations of four argument structure constructions (ASCs) and found similarities in behavior in the model and a sorting task done by humans.
However,  
\citet{zhou-etal-2024-constructions} found LMs failed to distinguish entailment differences between the causal excess construction (e.g., 'so heavy that it fell') and two structurally similar constructions  ('so happy that she won'; 'so certain that it rained'). 
}



\section{Conclusion}
\label{sec:conclusion}

Generalization is a key component of human language---and a big part of why LMs are successful at processing language. But we have argued that evaluations of the linguistic abilities of LMs are too often based on an assumption that generalization requires algebraic rules operating on words. Natural languages are not Lego sets. Instead, language involves flexible combinations of rich and varied constructions of differing sizes, complexities, and degrees of abstraction, which differ from algebraic rules in many ways. By designing new evaluations that accurately reflect the complexities of language, we can avoid under- or overestimating language models. The extent to which LMs produce and interpret combinations of constructions has only begun to be explored. We believe future progress lies not in asking whether LMs obey abstract rules, but in probing what kinds of constructions they learn, how they relate them, and how those structures guide novel interpretation and production. In doing so, we may better capture what it truly means to comprehend and use language.

\section*{Limitations}

While we have aimed to discuss benchmarks and evaluations in ways that reflect the historical trajectory as well as the present-day landscape, evaluations of LMs are continually developing. We feel the dominant paradigms have and continue to be based on data generated by rules and evaluated without regard for context effects, gradience, or semantic nuance, but we are keenly aware that we have likely overlooked metrics that go beyond rule-based evaluations \cite[e.g.,][]{parrish-etal-2021-nope}.  

We recognize the growing work in multilingual evaluations, which are inherently valuable \cite{mueller-etal-2020-cross,jumelet2025,kryvosheieva-levy-2025-controlled}. The current perspective applies to all natural languages, but comparative work is not the focus of the current perspective, and we use English examples for the sake of easy comprehension and brevity.


\section*{Acknowledgments}
We thank Najoung Kim, Kanishka Misra, and Will Merrill for helpful discussions and feedback. We are grateful to audiences at the NSF-sponsored New Horizons in Language Science workshop and the Analytical approaches to understanding neural networks summmer school
sponsored by Simon's Foundation for helpful feedback. Leonie Weissweiler was supported by a postdoctoral fellowship of the German Academic Exchange Service (DAAD).

\bibliography{anthology, literatur}

\end{document}